# Modern Machine and Deep Learning Systems as a way to achieve Man-Computer Symbiosis

**Chirag Gupta[1], BE Electronics and Communications**
[1]Thapar University, Patiala, PUNJAB 147004 INDIA

Corresponding author: Chirag Gupta (e-mail: chirag@cgupta.tech).

This work is an independent research done independently with no association to any institution.

**ABSTRACT** Man-Computer Symbiosis (MCS) was originally envisioned by the famous computer pioneer J.C.R. Licklider in 1960, as a logical evolution of the then inchoate relationship between computer and humans. In his paper, Licklider provided a set of criteria by which to judge if a Man-Computer System is a symbiotic one, and also provided some predictions about such systems in the near and far future. Since then, innovations in computer networks and the invention of Internet were major developments towards that end. However, with most systems based on conventional logical algorithms, many aspects of Licklider's MCS remained unfulfilled. This paper explores the extent to which modern machine learning systems in general, and deep learning ones in particular best exemplify MCS systems, and why they are the prime contenders to achieve a true Man Computer Symbiosis as described by Licklider in his original paper in the future. The case for deep learning is built by illustrating each point of the original criteria as well as the criteria laid by subsequent research into MCS systems, with specific examples and applications provided to strengthen the arguments. The efficacy of deep neural networks in achieving Artificial General Intelligence, which would be the perfect version of an MCS system is also explored.

**INDEX TERMS** Man Computer Symbiosis, Deep Learning, Machine learning, Human Computer Interaction

## I. INTRODUCTION

Man-Computer Symbiosis (MCS) was originally envisioned by J.C.R. Licklider in 1960, as a logical evolution of the then inchoate relationship between computer and humans. A significant development towards MCS was the growth of modern computer networks and the internet. Still, the most important facets covering both the central aims of the symbiosis as suggested by Licklider either remain unfulfilled or their presence is less clear.

It is significant to note that the original paper by Licklider was published in 1960, only 2 years after Rosenblatt's pioneering Perceptron paper [1], and almost contemporary to the invention of the first integrated circuit [2], and well before the invention of the programmable microprocessor [3]. Machine learning models like deep neural networks, and associated learning and optimization algorithms like gradient descent and backpropagation were not even conceived at that time, much less the efficacy of their application. Still, the concepts and language used by Licklider in his paper bears an uncanny resemblance to modern machine learning systems, both in basic research, and the applied ones which are being used to solve real world problems; all of which were developed in the decades that followed him.

While this undoubtedly shows the pioneering vision of Licklider himself, a task now falls on the current scientific community to ponder whether we have achieved true man-computer symbiosis, the role of machine learning and deep learning to that end, and what the future entails for such symbiotic relationships between man and machines. But first, we need to define what exactly is meant by a symbiotic relationship in the context of machines and computers.

### A. SYMBIOSIS AND THE CONTEMPRORY WORLD

A man-machine system is one in which both the human and machine interact closely, by means of which the operator performs a task based on a specified set of instructions [4]. Classical computer algorithms, and even most of the modern ones relying on precise logic are examples of man-machine systems. These conventional systems generally have either the human or the machine fully dependent on the other.

A symbiosis between man and machine is aptly described by the biological symbiosis, which is the mutually beneficial partnership between two organisms (host and symbiont) [5]. In his paper, Licklider claimed that while many man-machine systems were present at that time, a man-computer symbiosis remained elusive. Sixty years hence, after decades of development in computing hardware and software, some would argue that many such symbiosis already exist, mostly with the computer as host and the man playing the role of the symbiote [6]. An instant messaging application running on a smartphone may be cited as a modern example. A critical look however, would dismiss this particular claim, and other similar applications, as the human does not provide any benefit to the application, which in turn largely works exactly as programmed and designed by the developer; and hence cannot be a part of a mutually beneficial symbiotic relationship. Indeed, they resemble the 'mechanically extended man' as specified by North [7], and as such in such systems the human operator supplied the initiative, the direction, the integration, and the criterion.

As computational capability of machines has grown at a breathtaking pace [8], modern computer systems are so complex that they may create an illusion of a man-computer symbiosis already in operation. Indeed, some of the advanced systems resemble the kind of relationship envisioned by Licklider but they fall short on the two important criteria which must be met by a true symbiotic relationship between man and machine.

## II. MAN COMPUTER SYMBIOSIS

### A. CRITERIA FOR MAN-COMPUTER SYMBIOSIS

At this point, therefore, it would be necessary to define what we mean by a symbiotic relationship between man and computer. While a myriad of definitions is possible, a general rule is that there must be some mutual benefit to the computer as an equal part of the relationship. For this paper, we will take the original criteria which Licklider thought was appropriate then, which is:

1) The computer (machine) in the symbiosis must be capable of formulative thinking, as they are able to provide solutions to formulated problems based on a preprogrammed set of instructions. It follows that such a cognitive computer must be able to handle most if not all unforeseen alternatives (of the problem it is trying to solve). Licklider noted that if the user can think his problem through in advance, symbiotic association with a computing machine would not be necessary as conventional algorithms can be designed to handle those. Therefore, the computing machine would be capable of formulating ways to solve technical problems without explicitly being programmed to do so by a human.
2) The computer must effectively process the formulative thinking in 'real time', adapting to the changes in the variables relevant to the problem. It should be able to take varying data points of the dynamic environment surrounding it, observe patterns and relations between them, make inferences, and provide results in a form which is easily understood by the human with which it forms a symbiotic relationship.

Neal Lesh et al. further argued in their Man Computer Symbiosis Revisited paper [9] that true symbiotic interaction would require the following three elements in addition to the above two:

1) A complementary and effective division of labor between human and machine.
2) An explicit representation in the computer of the users' abilities, intentions, and beliefs.
3) The utilization of nonverbal communication modalities.

We will include all the above five points as our criteria to prove the efficacy of modern machine and deep learning systems as true MCS systems. Any system satisfying the criteria is a candidate for an MCS system. Not all systems or relationships satisfying the criteria would be MCS systems, but every true MCS system has to satisfy the criteria.

### B. CONVENTIONAL LOGICAL ALGORITHMIC SYSTEMS

With the frenetic pace of advancement in the electronic, computation, and software technology in the past two decades, humans have come to depend on programmed computers for a variety of generic and specific tasks in their daily lives [10]. While such unprecedented dependence on these non-sentient objects have engendered new ills [11], it can also be argued that this dependence and close relationship between humans and smart devices (smart in the conventional sense that the devices are able to take a lot of information from both the user and the surrounding environment, process them and provide results based on algorithms in their programming) must form an MCS system. After all, man is now more dependent on machines than any point in history, and the machines also are dependent on man in direct and indirect ways to perform their operation [12]. So, their relationship must be a symbiotic one.

While there is some credence in the above inference, but it is still unwarranted. We might say that the relationship between the computers of the present and the humans dependent on them form a symbiosis in a literal sense, it still falls short of a MCS system as according to our criteria. Primarily, these conventional logical systems, running on however advanced hardware or software, are not able to do formulative thinking. They are limited to providing solutions to formulated problems for which they have been programmed. Further, while modern computer systems react swiftly with changing data, they are still limited to the set for which they have been programmed. Consequently, their adaptation to changes in the input data or the problem itself is very limited and falls short of what must be the characteristic of an MCS system.

### C. DATA DRIVEN COGNITIVE MACHINE LEARNING SYSTEMS

It is now prudent to question if the modern machine learning models which are driven by data form a classical man-machine system, a man-computer symbiosis, or if they go a step further. As some of the machine learning techniques like deep neural networks are cognitive, in the sense that they can learn to perform a function with a reasonable fitness without explicit instructions of how to do it [13]. Therefore, such systems are more advanced than classical man-machine systems. But as of now, no deep neural network, however complex, can perform even a fraction of the range of varied tasks that a human brain is capable of [14]. A well trained deep neural network can perform only that task, either classification or regression, for which it has cognitively trained. A fully independent system which is able to learn new functions and adapt itself to changes in its environment on the fly, much like a human brain is in the realm of artificial general intelligence, which still remains an elusive goal [15]. Therefore, modern deep learning neural networks must lie somewhere in between. As these neural networks, like most of the other machine learning models are data driven and therefore rely on humans directly or indirectly to generate and provide raw data from which to learn, can be considered as candidates for a man-machine symbiotic system.

## III. DEEP LEARNING AS A WAY TO ACHIEVE MAN-COMPUTER SYMBIOSIS

Now after the exposition of all the prerequisites, we can deliberate on the hypothesis that 'Modern deep learning models and systems either already concur with man-computer symbiosis or such systems are most suited to achieve it in the near future'. For this, a comprehensive and critical analysis of the hypothesis, the supporting and opposing facts, current data and research, and future trends are required, while rigorously following the scientific method [16]. The same is done in the following sections.

### A. RATIONALE FOR CONSIDERING DEEP LEARNING IN PARTICULAR AS COMPARED TO MACHINE LEARNING IN GENERAL

To build a case for the hypothesis, this paper will now refer to a specific form of machine learning, that is Deep

Learning Neural Networks [17] and various forms of it from this point on. The reasons for taking this specific approach and not considering the entire or a large subset of the machine learning domain are as follows:

1) The domain of machine learning is vast and consists of a variety of models and algorithms. These range from simple linear regression models [18] to models of much higher complexity like transformer deep neural networks [19]. Simple linear models like linear regression fall more in the domain of statistics rather than that of computer science, and cannot be considered to form a MCS system for self-evident reasons. Therefore, the arguments made in favor of machine learning in general forming a MCS system would be incorrect, as those would not apply to all the different models in the machine learning domain equally. Therefore, prudence dictates to take specifically that model or algorithm which is most suited for a MCS system, and if a case is built for that, further generalizations can be drawn for the whole machine learning domain.

2) Among all the non linear machine learning models, deep learning neural networks are considered to be most generalizable, and in recent years have been at the center of most key advances in both basic and applied research in the machine learning domain [20], [21]. Many models and algorithms like Bidirectional Encoder Representations from Transformers (Bert) [22], Style GAN [23] etc. have achieved results which were till then considered to be the prerogative of the human brain. These reasons further warrant the selection of the deep neural network as the candidate for consideration of an MCS system.

## B. HOW DEEP LEARNING SATISFIES THE CRITERIA OF MAN-COMPUTER SYMBIOSIS

**Formulative Thinking**
A computer with formulative thinking would be capable of formulating ways to solve technical problems without explicitly being programmed to do so by a human. This does not mean to solve an unseen version of a problem which the computer has been programmed to solve. For example, a scientific calculator can solve any solvable differential equation of the second order, but it clearly is not capable of formulative thinking. A computer that is able to solve any problem, or is able to learn to solve a novel problem without being explicitly programmed to do so, would mimic a human brain. That computer would be cognitive and reactive, just like a human is able to learn from its surroundings, have intuition, respond to stimuli etc., and would be indistinguishable from an average human to other humans unless they are explicitly told so. This is an ideal symbiotic relationship of the highest level. Such a system is called an Artificial General Intelligence, and is discussed further in Future Developments. But for a valid MCS, much simpler systems may suffice.

A typical deep learning model also solves a specific type of a problem. But that solution is not programmed by a human, but rather the patterns which can be used to solve the problem are figured out by the neural network in its training phase. The common algorithms of learning i.e. gradient descent [24] and backpropagation [25] remain the same for all networks, be it a 3D CNN for Image segmentation [26] or a 100 layer LSTM RNN for generating text [27], [28]. Moreover, deep neural networks are capable of real time online learning through which they can adapt to changing environments if enough representative data is available [29], [30]. This real-time online learning is valid for both entirely software systems and hardware, robotic systems running on edge [31]. This combination of common learning algorithms, and real time learning form the basis of formulative thinking.

In the past 2 decades (2000-2020), many deep learning networks have shown this kind of formulative thinking. Examples include deep networks for Computer Vision problems like Image Segmentation [32], and for Natural Language Processing and Natural Language Generation like Bidirectional Transformers [22]. Details of such networks are discussed in the 'Evolution of Deep Learning and current state of the art' and 'Future Developments' section.

**Real-time and Adaptive Learning**
Hardly any other technology exemplifies this crucial aspect of MCS better than Deep Learning. Deep neural networks have a set of parameters representing the constants of many linear equations and some non linear equations. These parameters typically range from a few thousands to hundreds of millions [20]. Training a network essentially means to find a set of parameters, for which the error in training dataset is minimum. Each backward pass modifies the parameters by a small magnitude corresponding to the learning rate, which makes the whole network highly adaptive to data. There are many training and optimization algorithms specifically designed for this adaptive approach, with even the hyperparameters such as the learning rate being adaptive [33], [34]. As more or newer data becomes available, a deep neural network is able to undergo retraining on these new examples. This can be done in real-time even when the network is being used for inference, with 'online learning', and some neural network designs such as 'Deepwalk' [35] designed specifically for that purpose. It is also possible to take a network trained for some generic application and retrain just a part (generally the few ending layers) of the network to perform a specialized task using Transfer Learning techniques [36], [37].

**Division of Labor between Human and Machine**
In a man computer symbiotic system, the division of labor between the human and the machine must be complementary but clear and distinct. It should make the whole system more efficient in utilizing resources and producing results. In deep learning models, humans are ultimately responsible for collecting the data and preparing it in a form that can be accepted by the neural network. They are also responsible for the general configuration of the network and the extent to which the network is to be trained. The deep neural networks are responsible for identifying the patterns and optimizing the objective function (learning) by themselves using the optimization and fitting algorithms. This happens in multiple epochs using the data provided, with no manual intervention by the human [38]. Thus, the division of labor in training and running a deep learning system is both complementary and distinct.

**Representation of Users' Features and Utilization of Nonverbal Communication Modalities**
How deep neural networks abstract the training data and how they store the representations between their hidden layers is a matter of ongoing research. It can be verified, especially in convolutional neural networks (with 2D convolutional, Max Pooling pairs of layers) how the image representations of features at nodes get more abstract in each successive layer level of the network [39] (see figure 1). These feature representations are identified by the network itself during the training process. All the data to be used with the neural network, whether training, test, input, output must be in mathematical form. As long as the data or the features of a dataset can be represented in integer or floating-point vectors, they can be used with the neural network. Therefore, the deep neural network is a versatile machine and is able to communicate with humans and other machines in a variety of modes, both verbal and non verbal. For example, strings of text or lines of speech in natural language can be effectively exploited for a variety of intelligent tasks using deep learning for natural language processing [40]. At the same time these networks are equally effective at tasks like Computer Vision [41], [42], Drug development [43], [44], Healthcare and Medical research [45] etc. where the data and modes of communication is non verbal, and almost inscrutable to an average human. This versatility of deep learning to adapt to any data and for any application makes it a prime contender for an MCS system.

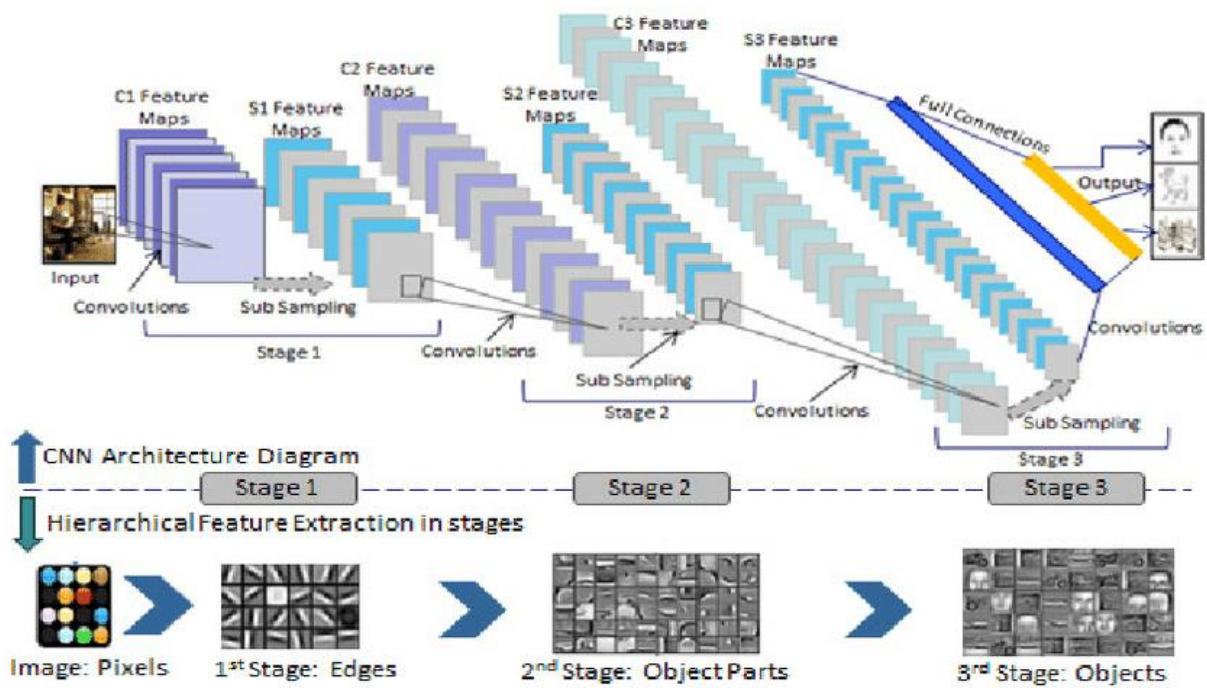

FIGURE 1.  Representative Image of layer level abstraction in a CNN [46]

## IV. FUTURE DEVELOPMENTS

Deep Learning systems are expected to continue to evolve into a more perfect version of Man-Computer Symbiosis (MCS). The immediate catalyst of their growing efficacy in solutions of many problems is due to the computational power of devices such as smartphones reaching a critical threshold (like the inclusion of dedicated neural units in the CPU die) [47] - [50] and novel networks (like MobileBERT) [51] which can extract significant value from data with less computation. It has been shown in this paper that current deep learning systems already exhibit the criteria specified for a true MCS system to varying degrees, and this conforming to it would continue to increase with time, as more research, both basic and applied is put into it.

Whether or not deep learning systems evolve into a true Artificial General Intelligence (AGI) [15] remains to be seen. AGI is the ultimate goal of intelligent machines which would form a perfect MCS or perhaps even transcend it. Research has shown the effectiveness of deep learning networks and deep reinforcement learning techniques as the foundation of AGI systems [52], [53]. However, at this stage it would be imprudent to conclude with certainty what deep learning systems would soon evolve into AGI as there is evidence and arguments both in favor and against it. At the same time, a good case has been built for machine and deep learning systems to become the Man-Computer Symbiotic systems as originally envisioned by J.C.R. Licklider. These systems interfaced with conventional algorithms and networking systems are closest to MCS in the present world, and consequently, the same systems are best suited to evolve into true MCS in the future, and ultimately be the catalyst for achieving Artificial General Intelligence.

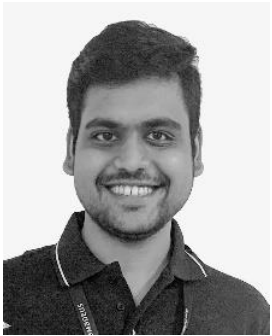

**CHIRAG GUPTA** was born in Jalandhar, Punjab, India in 1996. He received the B.E. degree in electronics and communications engineering from the Thapar University, Patiala, in 2018. He currently is part of the GCC R&D Innovation Labs team in Amadeus Labs, Bangalore working on applied NLP and deep learning to solve business problems. His research interests include machine learning, deep learning, natural language processing, and edge electronics.